\documentclass[11pt,letterpaper]{article}
\usepackage{emnlp2017}
\usepackage{times}
\usepackage{latexsym}

\usepackage{booktabs}
\usepackage{enumitem}
\usepackage[ruled,linesnumbered]{algorithm2e}
\usepackage{subfig}
\usepackage{graphicx}
\usepackage{mathtools}
\usepackage{amsmath,amssymb}
\usepackage{amsthm}
\usepackage{multirow}
\theoremstyle{definition}
\newtheorem{definition}{Definition}[section]

\usepackage{xcolor}

\usepackage{amsopn}
\DeclareMathOperator*{\argmax}{arg\,max}

\emnlpfinalcopy

\def\emnlppaperid{962}

\title{Learning Word Relatedness over Time}

\author{Guy D. Rosin\textsuperscript{1}, Eytan Adar\textsuperscript{2}, Kira Radinsky\textsuperscript{1,3} \\
         \textsuperscript{1}Technion -- Israel Institute of Technology, Haifa, Israel \\
         \textsuperscript{2}University of Michigan,  Ann Arbor, USA \\
         \textsuperscript{3}eBay Research, Israel \\
         {\tt \{guyrosin,kirar\}@cs.technion.ac.il, eadar@umich.edu}
}

\date{}

\begin{document}

\maketitle

\begin{abstract}
Search systems are often focused on providing relevant results for the ``now'', assuming both corpora and user needs that focus on the present. However, many corpora today reflect significant longitudinal collections ranging from 20 years of the Web to hundreds of years of digitized newspapers and books. Understanding the \textit{temporal intent} of the user and retrieving the most relevant historical content has become a significant challenge. Common search features, such as query expansion, leverage the relationship between terms but cannot function well across all times when relationships vary temporally. In this work, we introduce a temporal relationship model that is extracted from longitudinal data collections. The model supports the task of identifying, given two words, \textit{when} they relate to each other. We present an algorithmic framework for this task and show its application for the task of query expansion, achieving high gain.
\end{abstract}

\maketitle

\section{Introduction}
The focus of large-scale Web search engines is largely on providing the best access to present snapshots of text -- what we call the ``Now Web''. The system constraints and motivating use cases of traditional information retrieval (IR) systems, coupled with the relatively short history of the Web, has meant that little attention has been paid to how search engines will function when search must scale not only to the  number of documents but also temporally.  
Most IR systems assume fixed language models and lexicons. 
They focus only on the leading edge of query behavior (i.e., what does the user likely 
mean today when they type ``Jaguar'').  
In this context, features as basic as disambiguation and spelling 
corrections are fixed to what is most likely today or within the past few years \cite{Radinsky:2013:BDW}, query expansions and 
synonyms are weighted towards current information \cite{ShokouhiRadinsky:SIGIR:2012}, and results tend to include the most recent and popular content. 
While this problem would seem to be speculative in that it will be years until we need 
to address it, the reality is the rate of change \cite{Adar:2009:WSDM} of the Web, language, and culture have simply 
compressed the time in which critical changes happen.

The ``Now Web'' assumptions are entirely reasonable for temporally coherent text collections and allow users (and search engines) to ignore the complexity of changing language and concentrate on a 
narrower (though by no means simpler) set of issues.  The reality is that this serves a significant user 
population effectively.  There are nonetheless a growing number of both corpora and users who require access not just to 
what is relevant at a particular instant (e.g., Hathitrust~\cite{MEET:MEET14505001085}, historical news corpora, the Internet Archives, and even fast changing Twitter feeds). Within such contexts, a search engine will need to vary the way it 
functions (e.g., disambiguation) and interacts (e.g., suggested query expansions) depending on the 
period and temporal scale of documents being queried. This, of course, is further complicated by the 
fact that Web pages are constantly evolving and replaced.  

Take for example the query ``Prime Minister Ariel Sharon''. When fed into a news archive search engine, the likely intent was finding results about \textit{Sharon's} role as Israel's \textit{Prime Minister}, a role held from \textit{2001 to 2006}. \citet{Singh:2016:HDH:2854946.2854959} refer to this as a \textit{Historical Query Intent}. However, most popular search engines return results about Sharon's death in 2011, when he was no longer prime minister. The searcher may take the additional step of filtering results to a time period, but this requires knowing what that period should be. Other query features are also unresponsive to temporal context. For example, the top query suggestions for this query focus on more recent events of his death: ``Former prime minister Sharon dies at 85'', ``Former prime minister Sharon's condition worsens'', etc.  While these \textit{might} satisfy the searcher if they are looking for the latest results, or the results most covered by the press, there are clearly other possible needs~\cite{Bingham01062010}. 

In this paper, we focus on the task of measuring word relatedness over time. Specifically, we infer whether two words (tokens) relate to each other during a certain time range. This task is an essential building block of many temporal applications and we specifically target \textit{time-sensitive query expansion (QE)}. 
Our focus is on \textit{semantic relatedness} rather than \textit{semantic similarity}. Relatedness assumes many different kinds of specific
relations (e.g. meronymy, antonymy, functional association) and is often more useful for computational linguistics applications than the more narrow notion of  
similarity~\cite{Budanitsky:2006:EWM}. 
  
\begin{figure}
\centering

\subfloat[]{
  \includegraphics[height=5.5cm]{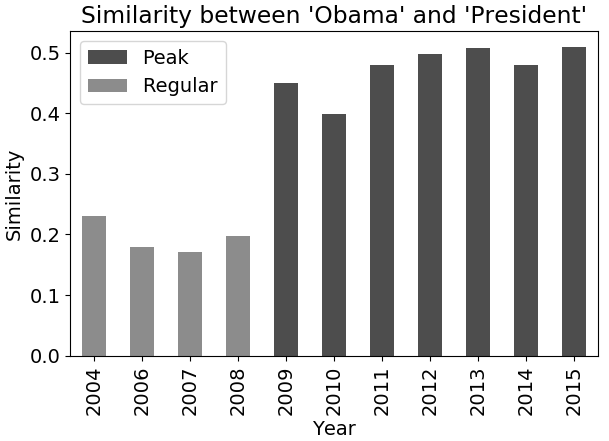}
  \label{fig:temp:obama}
}\qquad
\subfloat[]{
  \includegraphics[height=5.6cm]{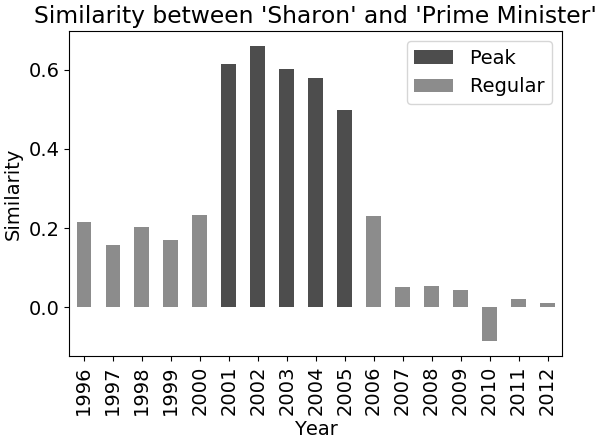}
  \label{fig:temp:sharon}
}

\caption{Similarity identified by our algorithms between words over time.
		Dark gray indicates high similarity whereas light gray indicates non-significant similarity.}
	\label{fig:obamapresident}
\end{figure}

We present several temporal word-relatedness algorithms. Our method utilizes a large scale temporal corpus spanning over 150 years (The New York Times archive) to generate \textit{temporal deep word embeddings}. We describe several algorithms to measure word relatedness over time using these temporal embeddings.
Figure~\ref{fig:temp:obama} presents the performance of one of  those algorithms on the words ``Obama'' and ``President''. Note that the highest relatedness score for the words appears during the presidential term of Barack Obama.
Similarly, Figure~\ref{fig:temp:sharon} shows a high score for ``Ariel Sharon'' and ``prime minister'' only during his term. 

\begin{figure}
	\includegraphics[scale=0.232]{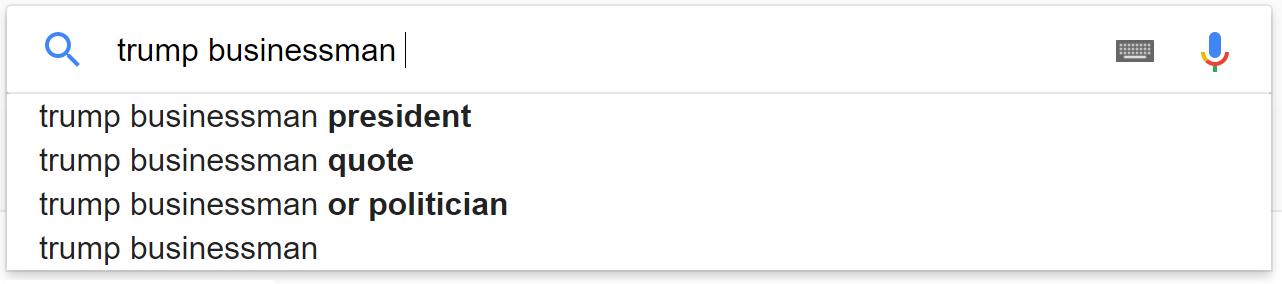}
	\caption{ Query expansions for the terms ``Trump Businessman''. Most results are referring to his term as president and not to his business activity.}
	\label{fig:trump}
\end{figure}

Using the approach above, we present a specific application -- producing temporally appropriate query-expansions. For example, consider the query: ``Trump Businessman''. Figure~\ref{fig:trump} shows the non-temporal query expansion suggestions which focus heavily on the first entity (i.e., ``Trump'') and his current ``state'' (i.e., a focus on Donald Trump as President, rather than presenting suggestions about Trump's business activity as implied by the query). We present an empirical analysis presenting the strengths and weaknesses of the different temporal query-expansion algorithms and comparing them to current word-embeddings-based QE algorithms  \cite{Kurland:2016:QE,Diaz:2016:QE}.

In this paper we describe a novel problem of evaluating word relatedness over time and contribute our datasets to evaluate this task to the community\footnote{\url{https://github.com/guyrosin/learning-word-relatedness}}.
Second, we present novel representations and algorithms for evaluating this task and show high performance. We share our code with the community as well.
Finally, we present the application of this task to query-expansion and present several methods built on top of the temporal relatedness algorithms that show high performance for QE.

\section{Related Work}
Understanding the semantic change of words has become an active research topic (Section~\ref{Section:WordDynamics}). Most work has focused on identifying semantic drifts and word meaning changes.
A lot of effort has been made into analyzing texts temporally: several methods for temporal information extraction were recently proposed \cite{ling2010temporal,kuzey2012extraction,talukdar2012coupled}, as well as publicly released knowledge bases, such as YAGO2 \cite{yago}. These methods automatically extract temporal relational facts from free text or semi-structured data.
In addition, \citet{pustejovsky2003timeml,uzzaman2012tempeval} and others annotated texts temporally, and extracted events as well as temporal expressions.

Work in Information Retrieval (IR, Section \ref{Section:TemporalSearch}) has discussed the concept of `time' as a contextual parameter for understanding user intent. Largely, this research utilizes query-log analysis with the time \textit{of the query} as a context signal. In this work, we leverage the temporal variation in word relatedness to understand, and better accommodate, intent.

\subsection{Word Dynamics}
\label{Section:WordDynamics}
Continuous word embeddings~\cite{word2vec} have been shown to effectively encapsulate relatedness between words.
\citet{Radinsky:2011:TSA} used temporal patterns of words from a large corpus for the task of word similarity. They showed that words that co-occur in history have a stronger relation. In our work, we focus on identifying \emph{when} a relation holds.
Numerous projects have studied the change of word meanings over time, and specifically focused on identification of the change itself. 
\citet{Sagi:2009:SDA} used Latent Semantic Analysis for detecting changes in word meaning. \citet{Wijaya:2011:USC} characterized 20 clusters to describe the nature of meaning change over time, whereas~\citet{Mitra:2014} used other clustering techniques to find changes in word senses. \citet{Mihalcea:2012} identified changes in word usage over time by the change in their related part-of-speech. Others have investigated the use of word frequency to identify epochs \cite{Popescu:2013}.

\citet{Jatowt:2014:FAS} represented a word embedding over the Google Books corpus (granularity of decades) and presented qualitative evaluation for several words. \citet{HamiltonJure:2016:ACL} built Word2Vec embedding models on the Google Books corpus to detect known word shifts over 30 words and presented a dozen new shifts from the data.
The authors presented two laws that govern the change of words -- frequent words change more slowly and polysemous words change more quickly. Finally, \citet{KenterAndRijke:2015:CIKM} studied changes in meaning (represented by a few seed words), and monitored the changing set of words that are used to denote it.

In our work, we focus on learning relatedness of words over time. We evaluate the technique using a large scale analysis showing its prediction accuracy.
Moreover, we define the task of identifying the temporality of relatedness between two words. We show that understanding the temporal behavior of entities improves performance on IR tasks.

\subsection{Temporal Search}
\label{Section:TemporalSearch}
The temporal aspects of queries and ranking gained significant attention in IR literature. Some have focused on characterizing query behavior over time. For example, different queries change in popularity over time~\cite{Wang:2003:MLW} and even by time of day~\cite{Beitzel:2004:hourlyanalysis}. \citet{Jones:temporalprofiles:2004} described queries to have three temporarilty patterns: atemporal, temporally unambiguous and temporally ambiguous. Others have leveraged temporal variance to look for indicators of query intent~\cite{Kulkarni:2011:understandingtemporal}. Several efforts \cite{Matias:2009:techreport,chien2005semantic,Vlachos:2004:identifyingsimilarities,Zhao:2006:time-dependentsemantic,Shokouhi:2011:SIGIR,
Radinsky:WWWPrediction:2012} were done to not only characterize the temporal query behavior but also model it via time-series analysis.
Radinsky and colleagues modeled changes in the frequency of clicked URLs, queries, and clicked query-URL pairs by using time-series analysis and show its application for improving ranking and query auto-suggestions~\cite{Radinsky:2013:BDW, ShokouhiRadinsky:SIGIR:2012}. \citet{Singh:2016:HDH:2854946.2854959} focused on serving the specific needs of historians, and introduced the notion of a \textit{Historical Query Intent} for this purpose.

Whereas prior work mainly focuses on query-log analysis and understanding the user's intent based on the time the query was issued or the document was changed, our work focuses on understanding the subtle changes of language over time that indicate a temporal intent. We show that understanding the temporal relatedness between words is a building block needed for understanding better query intent. Given two words or entities we identify when their relatedness was the strongest to help produce better query expansions. 


\section{Temporal Relatedness Dynamics and Semantics}
To address the task of understanding temporal relatedness, our approach consists of three main steps:
(1) Represent relatedness using word embeddings over time (Section~\ref{sec:representing_relations}).
(2) Model relatedness \textit{change} over time as a time series (Section~\ref{sec:dynamics}).
(3) Combine these to identify when relatedness relations hold temporally (Section~\ref{sec:learning_relatedness}).

\subsection{Representing Relatedness using Word Embeddings} \label{sec:representing_relations}
In our work we leverage the distributed representation framework of Word2Vec~\cite{word2vec} (specifically skip-grams).  Intuitively, given a word $w_t$, skip-grams attempt to predict surrounding words, e.g. $w_{t-2},w_{t-1},w_{t+1},w_{t+2}$ for a window size of $c=2$.

\begin{definition}
Let $C_i$ be the word context for a word $w_i$. In this work, we consider the context to be the surrounding words of a window size of $n$ 
$C_i = \{w_{i-n},w_{i-1},w_{i+1},w_{i+n}\}$.
We define $C^t_i$ to be the context for a word $w_i$ in time period $t$, i.e. only in the documents written during time $t$.
\end{definition}
\begin{definition}
We define the word context of a \textbf{specific embedding} of a year $y$ for a word $w_i$, to be $\{w_c \in C^y_i\}$. Intuitively, a specific embedding represents the embedding of a word in a certain year. We denote the vector representation of a word $w_i$ in a year $y$ by $v^y_i$.
\end{definition}
\begin{definition}
We define the word context of a \textbf{global embedding} for a word $w_i$ to be $C_i$.
Intuitively, a global embedding represents the embedding of a word over all time periods. We denote the vector representation of a word $w_i$ by $v_i$.
\end{definition}

Using Word2Vec, we approximate the semantic relatedness between two entities by the cosine similarity between their embeddings~\cite{turney2010frequency}.

\begin{definition}
A \textbf{temporal relation} $(e_1,e_2,t)$, where $e_i$ are entities and $t$ is a referenced time period, is said to be true if $e_1,e_2$ relate during $t$, i.e. their semantic relatedness during that time is relatively high.
For example: \textit{(Christopher Nolan, The Dark Knight, 2008)} is true, due to the fact that the movie, which was released in 2008, was directed by Nolan.
\end{definition}
\begin{definition}
The \textbf{dynamics} of two entities $e_1,e_2$ is defined to be the time series of the semantic relatedness of $e_1$ and $e_2$:
\begin{multline}
Dynamics(e_1,e_2) =  \\ \big\langle cos(v^{t_1}_1,v^{t_1}_2), \dots, cos(v^{t_n}_1,v^{t_n}_2) \big\rangle
\end{multline}
where $t_1,\dots,t_n$ are all the time periods.
\end{definition}

We model entities' relatedness change over time by constructing their \textit{dynamics}.
Recall Figure~\ref{fig:temp:obama}, which shows the semantic distance between the vector representations of \textit{Barack Obama} and \textit{President} over the years. Semantic distance is an accurate indicator of the time period when Obama was president. 
Therefore, our goal is to detect the time periods of high relatedness using peak detection.

\subsection{Understanding Relatedness Dynamics} \label{sec:dynamics}
When considering a relationship between entities, detecting time periods of high relatedness enables us to reveal and identify what we call ``periods of interest'' -- time periods in which the entities were the closest. In this section, we present an algorithm to identify these ``periods of interest''. 
Intuitively, when considering the \textit{dynamics} of two entities, its peaks represent the lowest distances over time. More formally, given two entities $e_1,e_2$ we want to construct their \textit{dynamics} and find peaks in it, i.e. the following sequence of time periods: $\big\{t_i \mid cos(v^{t_i}_1,v^{t_i}_2) \text{ is relatively high, and } t_i<t_j \text{ if }i<j\big\}$.

Traditional peak detection algorithms focus on detecting peaks that are sharp and isolated (i.e. not too many surrounding points have similar values)~\cite{peakdetection}. In our case, relations often imply continuous periods of peaks, e.g. Obama was president for eight years. Therefore, we need to detect peaks, as well as periods of continuous peaks (i.e., steps).

Let $L=(t_1,v_1),(t_2,v_2),\dots,(t_n.v_n)$ be a list of tuples, where $t_i$ are time periods and $v_i$ are values. Given $L$, the algorithm returns a list of peak periods, i.e. $\{t \in L \mid t \text{ contains a peak}\}$. 

The first step is to find relative maxima: we scan $L$ and look for pairs $(t_i,v_i)$ such that $v_{i-1}<v_i>v_{i+1}$, i.e. $v_i$ is a local maximum.
We apply two minimum thresholds in order to filter insignificant points:  An absolute threshold, which below it we consider the peak not to be significant enough, and a relative threshold which facilitates removing points that are much lower than the highest maximum.
The final step of the algorithm is to find plateaus which will also be considered part of the peak. For each peak point, the algorithm considers the point's surrounding neighbors. Using a threshold relative to the current peak, those identified as close in their value to the current peak are added to the peak list.

\subsection{Learning Temporal Relatedness} \label{sec:learning_relatedness}
We define the task of learning temporal relatedness: given two entities $e_1,e_2$, identify whether they relate to each other during a certain year $y$, i.e., whether the temporal relation $(e_1,e_2,y)$ is true. For example, in Figure \ref{fig:temp:obama}, Obama was related to President in 2010 but not in 2005. 

\subsubsection{Specific Classifier} \label{sec:relations_specific_svm}
The first method we present to classify word relatedness, employs a classifier that receives as inputs $v_i$ corresponding to the entities $e_i$ and an additional feature of the year. In our evaluation (Section \ref{sec:evaluation}) we will use about 40 years of data, i.e. the year feature will have 40 possible values.
The classifier will predict, given two entities, whether they relate to each other during a referenced year.

Let $Cl\colon \mathbb{R}^n \to \{0,1\}$ be a classifier mapping a vector of $n$ features $F=(f_1, \dots, f_n)$ to a label $L \in \{0,1\}$.
Let our feature vector be of the form:
\begin{equation}
F=(v^y_1 \| v^y_2 \| y)
\end{equation}
where $v^y_1$ is the first entity's specific embedding (i.e. at time $y$), $v^y_2$ is the second entity's specific embedding and $y$ is the year.
As a preliminary step, we need to train $Cl$ on a diverse dataset of positive and negative examples, i.e. true and false temporal relations. For this purpose, we utilize our temporal relations dataset (Section \ref{sec:relations_dataset}). From each relation, we extract a temporal relation $(e_1,e_2,y)$, calculate its feature vector and use it for training.

Given a new temporal relation, we apply $Cl$ to predict whether it is true, i.e. whether its entities relate during the referenced time. 

\subsubsection{Temporal Classifier} \label{sec:relations_temporal_svm}
Here we use a classifier similar to the one described in Section~\ref{sec:relations_specific_svm} (training is performed the same way), combined with input from the entities \textit{dynamics}.
First, we build the \textit{dynamics}, i.e. build specific word embeddings of $e_1$ and $e_2$ for every year $y$, and calculate their cosine similarity $cos(v^y_1,v^y_2)$. We then apply our peak detection algorithm (Section~\ref{sec:dynamics}) on the \textit{dynamics} and use its output as one of the classifier's features (denoted by \textit{isPeak}). As a result, our feature vector is:
\begin{equation}
F=(v^y_1 \| v^y_2 \| y \| isPeak)
\end{equation}

\subsection{Leveraging World Knowledge}
We apply our techniques to two corpora: the first is a temporal corpora (Section \ref{sec:nyt_dataset}), which we used for creating word embeddings, and the second is a relational corpora (Section \ref{sec:relations_dataset}), which we used for training our models and for evaluation.

\subsubsection{Temporal Corpora} \label{sec:nyt_dataset}
For constructing the corpora, we used The New York Times archive\footnote{\url{http://spiderbites.nytimes.com/}}, with articles from 1981 to 2016 (9GB of text in total).
Specific word embeddings were generated for every time period (i.e., year) using Word2Vec.
Each year's data was used to create word embeddings using Word2Vec's skip-gram with negative sampling approach~\cite{word2vec}, with the Gensim library~\cite{gensim}. We trained the models with the following parameters: window size of 5, learning rate of 0.05 and dimensionality of 140. We filtered out words with less than 30 occurrences during that year. 

We observe both \textit{ambiguity} (Apple, the company and the fruit) and \textit{variability} (different phrases referring to the same entity, e.g., \textit{President Obama}, \textit{Barack H. Obama}, \textit{Obama}). While such `noise' may be problematic, both the scale of the data and stylistic standards of The New York Times help. Additionally, we ensure a connection to an entity database (Wikipedia) and perform additional cleaning methods (lemmatization, removing stopwords).

\subsubsection{Relational Corpora} \label{sec:relations_dataset}
In this work, we use YAGO2~\cite{yago} as our relational corpora due to its temporal focus. The YAGO2 knowledge base contains millions of facts about entities, automatically extracted from Wikipedia and other sources. We use relations from YAGO2 to build our own dataset of temporal relations, which we use in all of our algorithms and evaluation -- as a source for temporal relations.

The dataset consists of temporal relations in the following format: \textit{($entity_1$, $entity_2$, year, type, class)}, where $entity_1$ and $entity_2$ are entities, $type$ is a relation type, and $class$ is true if the relation holds on $year$.
For example, \textit{(Tim Burton, Batman Returns, 1992, Directed, true)} and \textit{(Battle Mogadishu, Somalia, 2010, HappenedIn, true)}. 
\autoref{tab:relations_dataset_composition} shows the exact dataset composition. We built our dataset on all the relation types that have a temporal dimension in YAGO2: \textit{Directed, HoldsPoliticalPosition, IsMarriedTo, Produced, PlaysFor, HappenedIn}. The dataset contains 80K of such relations.

\begin{table}
	\centering
	\begin{tabular}{ccc}
		\toprule
		Relation Type & Count & \%\\
		\midrule
		Directed & 37713 & 47.42\\
		HoldsPoliticalPosition & 2765 & 3.48\\
		IsMarriedTo & 2210 & 2.78\\
		PlaysFor & 4376 & 5.50\\
		Produced & 23567 & 29.63\\
		HappenedIn & 8899 & 11.19\\
		\midrule
		Total & 79530 & 100\\
		\bottomrule
	\end{tabular}
	\caption{Relations Dataset Composition}
	\label{tab:relations_dataset_composition}
\end{table}

\section{Evaluation} \label{sec:evaluation}
\subsection{Experimental Methodology} \label{sec:relations_methods_compared}

We compare the methods described in Section~\ref{sec:learning_relatedness}, where for $Cl$ we chose to use a Support Vector Machine (SVM)\footnote{We used the implementation by the scikit-learn library~\cite{scikit-learn}.}, with an RBF kernel and C=1.0 (chosen empirically).
Two baselines were used for comparison. The first is the common non-temporal model, i.e. a classifier that uses the global (all-time) word embeddings and the following features: the two entities' global embeddings, and a year.
More formally,
\begin{equation}
F=(v_1 || v_2 || y)
\end{equation}
Given a new temporal relation, the classifier predicts whether it is true during the referenced year, and we output the classifier's prediction.
The second baseline we compare against is a standard text classifier that uses the global word embeddings as its only features, i.e. $F=(v_1 || v_2)$.

The dataset on which we perform the evaluation is described in Section~\ref{sec:dataset_construction}.
The dataset is not balanced: it contains more negative examples than positive ones. Therefore, for evaluating the methods that involve a classifier we use stratified 10-fold cross validation.
We remove relations from consideration if there is insufficient data in the corpora for that year (i.e., one of the entities was filtered out due to low incidence).

\subsection{Dataset Construction} \label{sec:dataset_construction}
Recall that our relational corpora consists of 80K temporal relations in the following format: \textit{($entity_1$, $entity_2$, year, type, class)}, where $type$ is a relation type, and $class$ is true if the relation holds on $year$.

For training and evaluating our classifiers we need negative examples as well as positive examples. We generate negative examples in the following way: for every relation in the corpora, we randomly sample 10 negative examples. We exclude the years of the true examples from the dataset's year range, and then randomly choose years for the negative examples. To illustrate, let us observe the case of \textit{Obama, President}: Obama was president from 2009-2016, so we sample negative examples from 1981 to 2008, such as \textit{(Obama, President, 1990, HoldsPoliticalPosition, false)}. The resulting dataset contains ~420K relations. We refer to it as the \textit{Temporal Relations Dataset}\footnote{\url{https://github.com/guyrosin/learning-word-relatedness}}.

\subsection{Main Results} \label{sec:results_relations}
\autoref{tab:relations_results} presents the results of our experiments.

\textbf{Baselines}: The Global and Global+Year baselines produced an AUC of 0.55 and 0.57, respectively. They both performed much worse compared to our methods, with F1 of 0.13.

\textbf{Specific Classifier} produced an AUC of 0.72. It has the highest recall score of all methods (0.88), but its other scores are relatively low.

\textbf{Temporal Classifier} produced an AUC of 0.83. As reported in \autoref{tab:relations_results}, it performed significantly better compared to all other methods, with $p<0.05$. We applied the Wilcoxon signed-rank test to calculate statistical significance.

\begin{table}
	\centering
	\begin{tabular}{p{1.7cm}lllll}
		\toprule
		Algorithm & Acc. & Rec. & Pr. & F1 & AUC\\
		\midrule
		Global & 0.67 & 0.26 & 0.08 & 0.13 & 0.57 \\
		Global+Year & 0.52 & 0.39 & 0.08 & 0.13 & 0.55 \\
		\midrule
		Specific & 0.52 & \textbf{0.91} & 0.32 & 0.47 & 0.73\\
		\textbf{Temporal} & \textbf{0.81} & 0.67 & \textbf{0.58} & \textbf{0.62} & \textbf{0.84}\\
		\bottomrule
	\end{tabular}
	\caption{Relatedness Learning Evaluation Results (Accuracy, Recall, Precision, F1, AUC)}
	\label{tab:relations_results}
\end{table}

\subsection{Performance Analysis} \label{sec:performance_analysis}
We tuned Word2Vec parameters by empirically testing on a random subset of our dataset: we set the vector size to be 140 and used a minimum threshold of 30 occurrences (per year). We found that this balanced the removal of noisy data while ensuring that key entities were retained. For constructing the Word2Vec models, the amount of data is crucial or it may lead to unreliable~\cite{Hellrich2016BadC} or inaccurate results. We saw a clear correlation between accuracy and number of occurrences of a participating word. That drove our decision to evaluate our algorithms only on New York Times articles from 1981 onwards -- where the number of articles per year is sufficiently large.

\section{Task Example: Query Expansion} \label{sec:qe}
Temporal relatedness learning can be used for various NLP and IR-related tasks. For example, it is a common practice in IR to expand user queries to improve retrieval performance~\cite{qe_survey}. Our technique allows us to produce temporally appropriate expansions. Specifically, given a query of $n$ entities $Q=\{e_1,e_2,\dots,e_n\}$, our task is to expand $Q$ with additional search terms to add to it to improve retrieval of relevant documents.

For example, consider the query ``Trump Businessman'' (Figure~\ref{fig:trump}). Current QE methods, which do not have a temporal aspect, focus on Donald Trump as President of the United States, a potentially erroneous result depending on the temporal focus of the searcher. A reasonable temporal expansion, might contain terms that relate to Donald Trump's business activity, such as ``billionaire'' or ``real estate''. Using our technique, the temporal focus of a query can be identified and appropriate expansions offered to the end-user.  Specifically, we can analyze the relationship between the query entities to identify the ``most relevant'' time period -- when those entities were strongly connected. Intuitively, the QE algorithms will identify the most relevant time period $t$ for the query entities, and find semantically related terms from that time, to expand the query with.

To tackle this task, we use the algorithms described in Section~\ref{sec:learning_relatedness}. Several different algorithms can utilize the temporal relation models for the task of query expansion. Let us introduce the following definitions for our QE algorithms:
\begin{definition}
	Let $NN_K^t(e)$ be the set of $K$ terms that are the closest to an entity $e$ in time $t$.
\end{definition}
\begin{definition}
	Let $NN_K(e)$ be the set of ``globally'' (all-time) closest terms to an entity $e$ .
\end{definition}
\begin{definition}
	\textbf{Mutual closeness} between an entity $x$ and a query $Q$ is defined by the sum of cosine similarities between $x$ and every $e \in Q$, i.e. 
\begin{equation}M_\mathit{cos}(x,Q) := \sum_{e \in Q} cos(x,e)
\end{equation}
\end{definition}

\subsection{Query Expansion Algorithms} \label{sec:qe_algorithms}
We describe alternative strategies to provide temporal query expansion ranging from a generic baseline to algorithms that leverage our embedding and classifiers. As a running example, we use the query: ``Steven Spielberg, Saving Private Ryan'' (Spielberg directed the movie in 1998). `Reasonable' (temporally relevant) expansions for this query might include: actors who played in this movie, other Spielberg films or similar films from the same time and genre, etc. 

\subsubsection{Baseline}
Following the results of ~\citet{qe:roy2016}, we consider a baseline method that expands each entity separately, based on global Word2Vec similarity. We define the set of candidate expansion terms as 
\begin{equation}
C = \bigcup_{e \in Q} NN_K(e)
\end{equation}
i.e., for each entity, we choose the closest $K$ global terms. For each $c \in C$, we compute the mutual closeness $M_\mathit{cos}(c,Q)$ and sort the terms in $C$ on the basis of this value. The top $K$ candidates are selected as the actual expansion terms.
The baseline (poorly) expands the query ``Steven Spielberg, Saving Private Ryan'' with ``Inglourious [Basterds], George Lucas''.  In some sense, one can see the relation -- both are war movies, and Lucas and Spielberg have worked together. However, Inglourious Basterds was created at 2009 and was directed by Quentin Tarantino.

\subsubsection{Globally-Based Classifier}
We use a heuristic and assume that the most relevant time period $t$ for the entities of the query is the time when the entities were the closest.
We use the classifier from our baseline method in Section~\ref{sec:relations_methods_compared}, whose goal is to estimate how relevant a year is to a given set of $n$ entities. Its features are the global word embeddings, as well as a year: $F=(v_1 \| v_2 \| \dots \| v_n \| y)$.

We apply the classifier to every year $y$, and choose the one with the highest returned probability of the true label as the most relevant time $t$.
\begin{equation}
t = \argmax_y\{Cl(v_1,v_2,\dots,v_n,y)\}
\end{equation}
We take as candidate-expansion terms the $K$ closest terms to each entity from that year, separately:
\begin{equation}
C = \bigcup_{e \in Q} NN_K^t(e)
\end{equation}
$C$ is then filtered as described in the baseline.

To train the classifier, for each temporal relation in our temporal relations dataset, we calculate its feature vector and use it for training.
Considering our example, this method wrongly chooses $t=2004$. In that year the entity \textit{Saving Private Ryan} does not exist, so we end up with a wrong expansion of ``Francis Ford Coppola film''.

\subsubsection{Temporal Classifier} \label{sec:qe_temporal_classifier}
As we have seen in the previous subsection, the globally-based classifier is limited in cases where time-specific knowledge might yield better results. Thus, in this method we use the specific classifier from Section~\ref{sec:relations_specific_svm}. Its features are the entities' specific embeddings, and a year: $F=(v^y_1 \| v^y_2 \| \dots \| v^y_n \| y)$.
We then continue as described in the previous method (find $t$, choose candidate terms and filter).

For our example, this method chooses correctly $t=1998$, which is exactly the year of \textit{Saving Private Ryan} release. Its expansion is ``Tom Hanks, Movie''. Since Tom Hanks had a lead role in the movie, the expansion is reasonable. The next algorithm produces the same expansion as well.

\subsubsection{Temporal Model Classifier}
This method uses the temporal classifier from Section~\ref{sec:relations_temporal_svm}. Its feature vector is: $F=(v^y_1 \| v^y_2 \| \dots \| v^y_n \| y \| \mathit{isPeak})$.
The rest is the same as described in Section~\ref{sec:qe_temporal_classifier}.

\subsection{Query Expansion Evaluation}
\textbf{Dataset.}
To evaluate temporal query expansion we use our temporal relations dataset, which will be made publicly available (described in Section~\ref{sec:relations_dataset}). First, we evaluate on queries consisting of two entities ($n=2$): for each relation, we create a distinct query that consists of its two entities concatenated. We search The New York Times corpus with this query\footnote{the archive was indexed using Solr (\url{https://lucene.apache.org/solr/})}.
We compare search performance when applying the various QE methods described in Section~\ref{sec:qe_algorithms}. To evaluate the methods that involve a classifier, we use stratified 10-fold cross validation, as the previous task was evaluated (Section~\ref{sec:relations_methods_compared}). We use $K=2$ for all methods, i.e. we generate two expansion terms per query.

In addition, we evaluate on queries consisting of three entities ($n=3$): we created a new dataset, which contains triplets of entities instead of pairs, by merging every two related (true) relations from our relations dataset. Two relations are considered related if they share an entity, and their time periods overlap. We then generate negative relations as described in Section~\ref{sec:dataset_construction}. In this new dataset, each temporal relation consists of three entities, a year and a binary classification.

\textbf{Evaluation Metrics.}
Though a complete evaluation of QE is beyond the scope of this paper we describe here an evaluation suited for the temporal case. It should be noted that the technique we propose here would likely be used alongside established QE techniques (e.g., log mining).

First, when providing query expansions and suggestions we would like for them to not only retrieve relevant content, but \textit{temporally}-relevant content. To test the latter we say that given a temporal relation, a retrieved article is considered ``true'' if it were published within the referenced time, and ``false'' otherwise. Additional manual validation was done to evaluate its relevance to the query.
This metric, while not the most accurate one, allows us to distinguish between results from the most relevant time period and others. Precision of the top 10 retrieved documents (P@10) is used to evaluate the retrieval effectiveness.

\textbf{Results.}
The results of the QE evaluation are reported in \autoref{tab:qe_results}. We observe a consistent behavior for different query sizes ($n=2,3$). For our temporal classifiers, for $n=3$ there is a 30\% increase in precision, compared to $n=2$.

All of our methods performed significantly better compared to the baseline (statistical significance testing has been performed using paired t-test with $p<0.05$). This establishes our claim that utilizing temporal knowledge yields more temporal--promising results.
The \textit{Temporal Model Classifier} showed the best performance of all. This, too, suits our claim and fits to the results from the previous task (Section~\ref{sec:results_relations}).

\begin{table}
	\centering
	\begin{tabular}{ccc}
		\toprule
		\multirow{2}{*}{\raisebox{-\heavyrulewidth}{Method}} & \multicolumn{2}{c}{P@10} \\
  \cmidrule{2-3}
  & $n=2$ & $n=3$ \\
		\midrule
		Baseline \cite{qe:roy2016} & 14.0\% & 17.7\%\\
		Globally-Based Classifier & 18.0\% & 22.7\%\\
		Temporal Classifier & 27.1\% & 38.5\%\\
		Temporal Model Classifier & \textbf{29.4\%} & \textbf{39\%}\\
		\bottomrule
	\end{tabular}
	\caption{Results of QE Algorithms Evaluation}
	\label{tab:qe_results}
\end{table}

\subsection{Textual Relevance} \label{sec:discussion}
To validate results, we compared our query expansion algorithms' performance on different relation types and found big differences. On the relations \textit{HoldsPoliticalPosition, HappenedIn} and \textit{IsMarriedTo}, the temporal algorithms achieved around 50\% accuracy, while on \textit{Directed} and \textit{Produced} they got only 20\%. This difference is reasonable, as our models were built upon a news corpus.

Let us observe an example of using the QE algorithms with the query ``Vicente Fox President'' (Fox was president of Mexico from 2000 to 2006). The baseline expands with Mexico's two previous presidents (Zedillo and Salinas). This makes sense as the baseline doesn't take time into account.
The globally-based classifier expands with Roh Moo-hyun, who was president of Korea during the same time period.
\textit{Temporal Classifier} expands with ``Ricardo Lagos, National Action Party'' (Lagos was president of Chile during that time. The latter is Fox's political party).
The \textit{Temporal Model Classifier} expands with `presidential' and Francisco Labastida (the candidate who lost the elections to Fox).

\begin{figure}
\centering
	\includegraphics[width=0.48\textwidth]{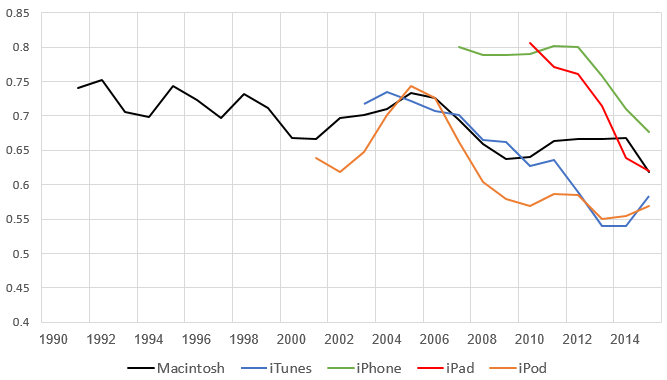}
	\caption{Similarity between Apple and its top products over time. The y-axis is cosine similarity.}
	\label{fig:apple_graph}
\end{figure}

Figure~\ref{fig:apple_graph} shows the similarity between Apple and its top products since 1990. We can infer which products were the most significant at each time. Take as example the query ``Apple Steve Jobs''. Using our technique, we can find the most relevant time period for this query, which is from Apple's foundation in 1976 until Jobs' death in 2011. Leveraging Figure~\ref{fig:apple_graph}, we can expand this query focusing on the most popular products of that time.

\section{Conclusions}
We believe that as corpora evolve to include temporally-varying datasets, new techniques must be devised to support traditional and new IR methods. In this paper, we introduced a novel technique for extracting relations in temporal datasets. The technique is efficient at large scales and works in an unsupervised manner.  Our experiments demonstrate the viability of the extraction technique as well as describing ways that it can be used in downstream applications. We specifically demonstrate a number of query expansion algorithms that can benefit from this technique. 



\bibliographystyle{emnlp_natbib}
\bibliography{sigproc,AC,webmodeling}

\end{document}